\pdfoutput=1
\documentclass[11pt]{article}

\usepackage[]{acl}

\usepackage{times}
\usepackage{latexsym}

\usepackage[T1]{fontenc}
\usepackage[utf8]{inputenc}

\usepackage{microtype}

\usepackage{epsfig}
\usepackage{amsmath}
\usepackage{amssymb}
\usepackage{microtype}
\usepackage{xspace}
\usepackage{subfig}
\usepackage{graphicx,wrapfig}
\usepackage{booktabs}
\usepackage{makecell}
\usepackage{framed}
\usepackage{multirow}
\usepackage[linesnumbered,algoruled,boxed,noend]{algorithm2e}
\usepackage{enumitem}

\usepackage{pifont}

\newcolumntype{L}[1]{>{\raggedright\let\newline\\\arraybackslash\hspace{0pt}}m{#1}}
\newcolumntype{C}[1]{>{\centering\let\newline\\\arraybackslash\hspace{0pt}}m{#1}}
\newcolumntype{R}[1]{>{\raggedleft\let\newline\\\arraybackslash\hspace{0pt}}m{#1}}

\SetKwInput{KwInput}{Input}
\SetKwInput{KwOutput}{Output}
\SetEndCharOfAlgoLine{}

\newcommand{\para}[1]{\smallskip\noindent\textbf{#1}}

\title{Leveraging Visual Knowledge in Language Tasks: An Empirical Study on Intermediate Pre-training for Cross-modal Knowledge Transfer}
\author{
Woojeong Jin\textsuperscript{1}\thanks{~~{Authors contributed equally.}},~
Dong-Ho Lee\textsuperscript{1}$^*$,~
Chenguang Zhu\textsuperscript{2},~
Jay Pujara\textsuperscript{1},~
Xiang Ren\textsuperscript{1}
\\
\textsuperscript{1}Department of Computer Science, University of Southern California\\
\textsuperscript{2}Microsoft Cognitive Services Research Group\\
{\small \texttt{\{woojeong.jin,dongho.lee,jpujara,xiangren\}@usc.edu} \texttt{\{chezhu\}@microsoft.com}}\\
}

\begin{document}
\maketitle
\begin{abstract}
%Can we leverage visual knowledge in improving language models on concept-centric NLP tasks?
% Pre-trained Language models have achieved impressive improvements on diverse natural language understanding (NLU) tasks including question answering, commonsense reasoning, etc.
Pre-trained language models are still far from human performance in tasks that need understanding of properties (e.g. appearance, measurable quantity) and affordances of everyday objects in the real world since the text lacks such information due to reporting bias.
In this work, we study whether integrating visual knowledge into a language model can fill the gap.
We investigate two types of knowledge transfer: (1) \textit{text knowledge transfer} using image captions that may contain enriched visual knowledge and (2) \textit{cross-modal knowledge transfer} using both images and captions with vision-language training objectives.
On 5 downstream tasks that may need visual knowledge to solve the problem, we perform extensive empirical comparisons over the presented objectives.
Our experiments show that visual knowledge transfer can improve performance in both low-resource and fully supervised settings.
\footnote{\href{https://github.com/INK-USC/CMKT}{https://github.com/INK-USC/CMKT}}
% We investigate multiple intermediate pre-training objectives on 
% %text corpora which models learn from lack knowledge that are not described in text due to the reporting bias.
% %In this work, we study the effect of visual knowledge in language learning. 
% We utilize image-caption data to transfer knowledge from visual domains to language models in pretraining and test the pretrained language models on various downstream tasks.
\end{abstract}

\section{Introduction}
% \xiang{\textbf{1st para (motivation)}: text LM got good performance on a range of language tasks; but still gaps to human-level performance. One reason for the gap is the lack of knowledge that are not represented in web text and encyclopedia text. For example visual  knowledge about common objects, their attributes and affordance, and their relationship may not be explicitly described in text because xxx. It is thus important to study the problem of how to leverage visual knowledge from external sources to help LMs on language tasks. 
% }

%%% terminology throughout the whole paper

Pre-trained language models (PTLMs) such as BERT~\cite{devlin2018bert}, RoBERTa~\cite{liu2019roberta}, and T5~\cite{2020t5} have shown impressive results in various conventional natural language understanding (NLU) tasks by capturing syntactic and semantic knowledge from the pre-training tasks of \textit{masked language modeling} and \textit{masked span infilling} tasks on massive text corpora.

\begin{figure}[t!]
    \centering 
    \includegraphics[width=0.98\columnwidth]{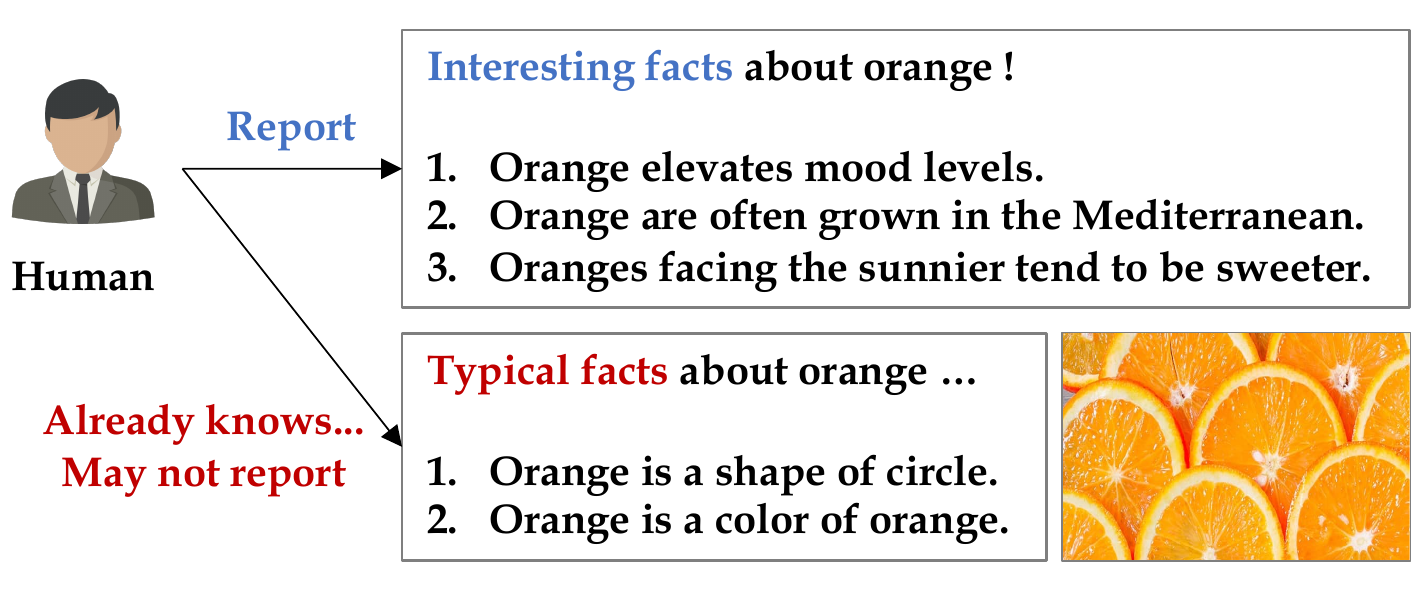}
    \caption{\textbf{Reporting Bias.} People tend to report what interests them rather than typical and general facts.}
    \label{fig:reporting_bias}
    \vspace{-0.4cm}
\end{figure}

Though yielding good performance on various NLU downstream tasks, these pre-training objectives suffer from a lack of out-of-domain knowledge that is not explicitly present in the pre-training corpus~\cite{gururangan-etal-2020-dont,petroni-etal-2021-kilt,DBLP:conf/aaai/SchickS20}.
Specifically, one type of knowledge that models often struggle with is the visual knowledge of common objects such as attributes (e.g. appearance, measurable quantity) and affordances. 
This is because this kind of knowledge  is rarely explicitly described in the training text due to reporting bias.
For example, as shown in Figure~\ref{fig:reporting_bias}, people tend to report what interests them rather than general facts such as a shape or color of oranges they already know.

% \xiang{\textbf{2nd para (related work)}: couple lines of work are related to addressing this gap: 
% 1) injecting KG info to the LMs (cite some papers) -- but the KG coverage, esp on visual knowledge, is a bottleneck;
% 2) retrieval-augmented and generation-augmented methods (cite, such as Peifeng's PathGen and Bill's KagNet) -- but it also require identifying a corpus to use, which may not represent visual knowledge well due to reporting biases. 
% 3) Intuitively, we want to equip the text LM encoders with more visual knowledge from the visual data like images. Briefly mention some pilot work like VidLanKD and say clearly about its limitations.
% --> In this work, we look to systematically analyze and understand whether we can leverage such visual knowledge by doing intermediate pretraining with different knowledge sources.
% }
% To fill the gap between reality and text learned by PTLMs with more knowledge, 
Towards better knowledge-enhanced PTLMs, recent works incorporate external knowledge bases (e.g., knowledge graph, dictionary) to inject entity knowledge into PTLMs~\cite{zhang-etal-2019-ernie, peters-etal-2019-knowledge, wang-etal-2021-k, yu2021dict} or retrieve knowledge from external knowledge bases to solve the problem~\cite{lin-etal-2019-kagnet, wang-etal-2020-connecting}. 
However, these approaches still suffer from a lack of visual knowledge that is important to understand the real world.

In this paper, we conduct systematic experiments to understand whether such visual knowledge can be transferred into LMs, and if so, how to perform effective knowledge transfer. Specifically, we look into a series of analysis question as follows:
(1) Can intermediate pre-training~\cite{pruksachatkun-etal-2020-intermediate} on image-caption pairs help transfer the knowledge? 
(2) What types of knowledge sources are more helpful? 
To answer questions, we explore various intermediate pre-training tasks~\cite{pruksachatkun-etal-2020-intermediate} on two different sources: text-only (\textit{text knowledge transfer} from visual domains) and image-caption pairs (\textit{cross-modal knowledge transfer}).

% In this paper, we conduct a series of experiments and systematic analysis to understand whether such visual knowledge can be transferred into LMs.
% To inject visual knowledge into LMs, we explore various intermediate pre-training tasks~\cite{pruksachatkun-etal-2020-intermediate} on two different sources: text-only (\textit{text knowledge transfer} from visual domains) and image-caption pairs (\textit{cross-modal knowledge transfer}).
% \xiang{Missing: 
% (1) 
% (2) Summarize the (high-level) analysis questions we have when conducting the study.
% }

%through \textit{intermediate pre-training} on various knowledge sources including text and images.

\begin{figure*}[t!]
    \centering 
    \includegraphics[width=\textwidth]{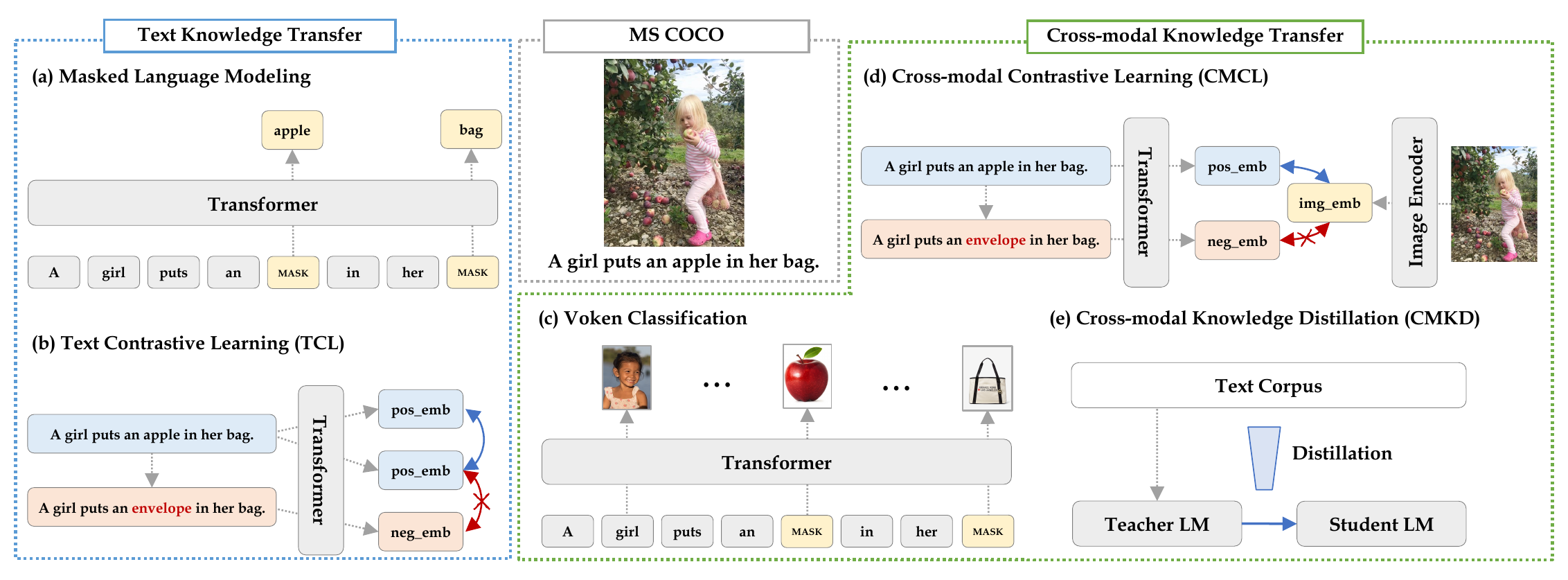}
    \caption{\textbf{Illustration of different methods for transferring visual knowledge into transformer-based language model.} In this example, we assume image-caption pair as an input. 
    (a) \textit{masked language model} \cite{devlin2018bert} on image captions. 
    (b) \textit{text contrastive learning} obtains positive example by dropout representation to learn better sentence representation while negative augmentation is optional. 
    (c) \textit{voken classification} employs token-level text-to-image retrieval to transfer visual knowledge. 
    (d) \textit{cross-modal contrastive learning} aims to train correct paring of images and captions. 
    (e) \textit{cross-modal knowledge distillation} transfers knowledge from the teacher model, which is trained by cross-modal contrastive learning, into student model. 
    % \wj{img encoder, knowledge source, mlm, voken, e. vlbert,vokenization, blue arrow confusing.}
    % \xiang{1) Stress more on the knowledge sources/corpora used in the \textbf{intermediate pretraining} --- since that's also the important aspect of why we can call this knowledge transfer; 2) clarify the de-coupled model for the CM-CL/KD case.}
    }
    \label{fig:overview}
    \vspace{-0.2cm}
\end{figure*}

% \xiang{\textbf{4th para (baselines and methods for comparison)}: we introduce in more details about the baselines and variants/design choices we explore, from a couple dimensions (data sources, obj designs, negative sample designs), 
% }
For the text knowledge transfer, we utilize text corpus from visual domain, e.g., image captions.
We leverage two training objectives for the language model:
(1) \textit{masked language modeling} follows the domain adaptive pre-training scheme~\cite{gururangan-etal-2020-dont}, assuming the corpus contains enriched visual knowledge or physical commonsense knowledge;
(2) \textit{text contrastive learning} augments the sentence representation with dropout to create positive samples while considering all others in the batch as negative samples for the contrastive learning~\cite{gao2021simcse}, assuming training better sentence representations leads to better understanding of the corpus.

% \xiang{1) Update to include VLBERT; 2) speak out more about the ANS and PSA variants we explore, since it leads to interesting results.}
For the cross-modal knowledge transfer, we explore multiple methods to transfer visual-related knowledge to LMs:
(1) \textit{masked language modeling with visual clues} incorporates visual clues to capture dependencies between visual and linguistic contents~\cite{su2019vl}; 
(2) \textit{voken classification} contextually aligns language tokens to their related images (called "vokens") to transfer visual knowledge into LMs~\cite{tan2020vokenization};
(3) \textit{cross-modal contrastive learning} aims to improve text representations by maximizing the agreement between correct image-text pairs versus random (in-batch) and adversarial negative pairs by contrastive learning between image and text modalities; and
(4) \textit{cross-modal knowledge distillation} transfers the knowledge from the teacher model, which is trained by cross-modal contrastive learning on image and text modalities, to the student language model using knowledge distillation.
We perform comprehensive comparisons on five downstream tasks that may require visual or physical commonsense knowledge, including PIQA~\cite{bisk2020piqa}, Visual Paraphrasing (VP)~\cite{lin2015don}, CSQA~\cite{talmor2018commonsenseqa}, OBQA~\cite{mihaylov2018can}, and RiddleSense~\cite{lin2021riddlesense}. Results suggest that:
(1) Simple intermediate pre-training on captions can help improving performance on commonsense reasoning that needs physical or visual knowledge.
(2) Cross-modal knowledge transfer approaches consistently improve the performance in a large margin when only few train examples are available.
(3) Cross-modal contrastive learning shows that it is best for packaging visual knowledge into LMs.

% To summarize, our contributions include:
% (1) We present the systematic study for transferring visual knowledge into LMs. Five objectives on two different sources are investigated.
% (2) We perform comprehensive comparisons on five downstream tasks that may require visual or physical commonsense knowledge:PIQA~\cite{bisk2020piqa}, Visual Paraphrasing (VP)~\cite{lin2015don}, CSQA~\cite{talmor2018commonsenseqa}, OBQA~\cite{mihaylov2018can}, and RiddleSense~\cite{lin2021riddlesense}.
% (3) Compared to existing approaches, the proposed approaches achieve better performance.

\section{Analysis Setup}
In this work, we study how to transfer the visual knowledge into language models.
For this study, we introduce our analysis setup: problem formulation, analysis questions, and knowledge corpora.

% \xiang{3-4 sentences to overview the motivation and analysis scope.}

% \xiang{We need two illustrative figures: 1) a double-column one to illustrate side-by-side different methods in terms of their methodology and knowledge sources;
% 2) A more detailed illustration on the CMCL method variants and CMKD method.}

% Throughout this section, we assume the data is a collection of image-caption pairs $\left\{(x_i^v, x_i^l)\right\}_{i=1}^{m}$ and image encoder $f_V$ and text encoder $f_L$ are given.

\subsection{Problem Formulation}
We focus on a pre-trained text encoder $f_L$ and an image encoder $f_V$ if images are available.
$f_L$ and $f_V$ are initialized with pre-trained model and we continue to pre-train the models on different sources and tasks, which we call \textit{intermediate pre-training}~\cite{gururangan2020don,pruksachatkun2020intermediate}.
% Here, we freeze the $f_V$ and train a final layer of it to let the text encoder learn more information from intermediate pre-training. 
After the intermediate pre-training, we fine-tune $f_L$ on downstream NLU tasks.
Existing NLU benchmarks have been trained against standard supervised learning paradigms that typically require a large number of question answering examples which need a large annotation efforts.
However, in scenarios where the number of labeled examples is small, the model tends to overfit the training examples and shows poor generalization performance on test set.
Here, we evaluate the intermediate pre-training objective's generalization ability on test set in both fully supervised and low-resource settings.

% \xiang{
% % This is an important part where we introduce the workflow of the study:
% % 1) our focus is the masked LMs for language tasks, like BERT and Roberta. Our starting point is the pretrained LMs.  (define the necessary notations throughout this subsection)
% % 2) !!goal: we look to improve the LM by transferring visual knowledge about common objects from image caption data to the LM
% 1) A detailed description of the workflow: intermediate pretraining on what; then finetune on what.
% 2) what are the measure of success --- i.e., evaluation aspects.
% }

% \subsection{Knowledge Corpus}
% \para{Text corpus.}
% Wiki103~\cite{merity2016pointer}, BooksCorpus~\cite{}, ConceptNet~\cite{}

% \xiang{this is for introducing the kinds of corpora we used in the study. Let's category them into buckets based on the modality and kinds of knowledge, and introduce to some level of details.}

% \para{Image-caption dataset.}
% We collect image-caption data, MS COCO~\cite{lin2014microsoft,chen2015microsoft}, to do knowledge transfer.
% \xiang{have a table to give stats of these corpora too.}

\subsection{Analysis Questions}
% \xiang{Need more expansion on the evaluation procedure and protocol (measure of the outcome/success) for each analysis question.}
In this paper, we provide a comprehensive study for transferring the visual knowledge into LMs.
Visual knowledge transfer can be done in two approaches, depending on the source to be trained:
(1) \textit{Text knowledge transfer} using the text corpus in the visual domain, e.g., image captions and
(2) \textit{cross-modal knowledge transfer} which passes visual knowledge about common objects to LMs by training over paired image and captions.
By evaluating the model on 5 downstream datasets that require physical and visual commonsense knowledge, we explore following three research questions.

% (1) text knowledg
% If we use only the text corpus that may contain bunch of 
% In this paper, we provide a comprehensive study for transferring the visual knowledge into LMs, and explore three main research questions.

% We conduct experiments to answer the following analysis questions using different evaluation setups.
% \xiang{Expand more on what things we're exploring here at a high level: knowledge corpus; downstream datasets; methods for intermediate pretraining; evaluation settings, etc.}

% \xiang{
% A more general framing of the analysis questions: 
% Q1: Can intermediate pretraining on external knowledge sources help transfer visual knowledge to augment text encoders?;
% Q2: What types of knowledge sources are more helpful for visual knowledge transfer?;
% Q3: What intermediate pretraining objectives are effective for cross-modal knowledge transfer?
% Q4: How can we guide the intermediate pretraining to focus on most challenging cases for the text encoders?

\para{Q1: Can intermediate pre-training on external knowledge sources help transfer visual knowledge to augment text encoders?}
% \wj{can we do that? does that really helpful corpus? img-caption data?}
% \wj{caption, img-caption (real image), input pairs (x,y)-> y, and x,y}
% We continually pre-train language models on captions of image-caption datasets. Here, we assume captions include enriched visual description of real world.
% To answer this question, we pre-train language models using caption datasets such as MS COCO~\cite{lin2014microsoft,chen2015microsoft}.
% , ConceptNet~\cite{}, Wiki103~\cite{merity2016pointer}, and BooksCorpus~\cite{}.
We investigate diverse intermediate pre-training methods with external knowledge sources including caption data to inject visual information from images and captions into LMs.
We first analyze the performance of text and cross-modal knowledge transfer methods with a image-caption dataset, and we additionally study text knowledge transfer methods with other text corpora such as GenericsKB~\cite{Bhakthavatsalam2020GenericsKBAK}, Wiki103~\cite{merity2016pointer} and BookCorpus~\cite{Zhu_2015_ICCV}.

% We investigate two different pre-training objectives (masked language modeling, contrastive learning) to inject visual information from captions into LMs.
% To explore whether the caption is best suited for packing visual knowledge into LM, we explore other text corpora such as GenericsKB~\cite{Bhakthavatsalam2020GenericsKBAK}, Wiki103~\cite{merity2016pointer} and BookCorpus~\cite{Zhu_2015_ICCV}.

\para{Q2: What types of knowledge sources are more helpful for visual knowledge transfer?}
As mentioned above, we have two categories to exploit visual information: (1) \textit{text knowledge transfer} and (2) \textit{cross-modal knowledge transfer}.
Here, we explore which type of knowledge transfer is more useful to transfer the visual knowledge into LMs.

% which exploits image representations to transfer the visual knowledge into LM.
% Here, we explore two options and analyze which sources are useful to transfer the visual knowledge into LM.
% Caption datasets contain two different sources, images and captions. We examine the sources by pre-training LMs with caption-only or image-caption sources.
%\xiang{Q3: How this can be done effectively?}
\para{Q3: What intermediate pre-training objectives are effective for cross-modal knowledge transfer?}
%\wj{how this can be done? architecture, different methods, Q3,4 merge into one question.}
We present three pre-training objectives for cross-modal knowledge transfer:
(1) voken classification,
(2) contrastive learning, and
(3) knowledge distillation.
Here, we want to present which strategy is best suited for cross-modal knowledge transfer.
Furthermore, we study how to enhance cross-modal contrastive learning with adversarial negative samplings.
% To answer this question, we explore pretraining objectives including masked language modeling, knowledge distillation, and contrastive learning.
% How can we guide the intermediate pretraining to focus on most challenging cases for the text encoders?
%\para{Q4: How can we guide the intermediate pretraining to focus on most challenging cases for the text encoders?}

\subsection{Pre-training Data}
To transfer the visual knowledge, we collect  250K image-caption pairs from MS COCO~\cite{lin2014microsoft,chen2015microsoft}.
MS COCO contains images reflecting the composition of actual everyday scenes and corresponding captions which describe contextual reasoning between objects in the scene.
We only use captions for text knowledge transfer while we use both images and captions for cross-modal knowledge transfer.
As an ablation study, we explore other text corpora such as GenericsKB~\cite{Bhakthavatsalam2020GenericsKBAK}, Wiki103~\cite{merity2016pointer} and BookCorpus~\cite{Zhu_2015_ICCV}.
% \xiang{Need significant expansion for this part, since it is core to our analysis:
% 1) No mention of the text corpora??
% 2) Separate the introduction on text corpora and img-caption data.
% 3）Say why we pick these corpora; 
% 4) what knowledge we believe they carry that can complement/enrich the text encoders; what are the main genre and characteristics of the data.
% }

\begin{table}[!t]
	\centering
	\small
	\resizebox{0.98\columnwidth}{!}{
		\begin{tabular}{lcccc}
            \toprule
            \textbf{Dataset} & $\#$ Train & $\#$ Dev & $\#$ Test & $\#$ choices \\
            \midrule
            PIQA & 14,113 & 1,838 & 2,000 & 2 \\
            VP & 21,988 & 2,000 & 6,057 & 2 \\
            CSQA & 8,500 & 1,221 & 1,241 & 5 \\
            OBQA & 4,957 & 500 & 500 & 4 \\
            RiddleSense & 3,510 & 1,021 & 1,202 & 5 \\
            \bottomrule
        \end{tabular}
	}
	\caption{\textbf{Downstream task data statistics.} We create in-house test set for PIQA and CSQA, and in-house dev set for VP by splitting the train set.
	}
	\label{tab:dataset}
\end{table}

\subsection{Downstream Tasks and Datasets}

For downstream benchmarks, we find tasks that can benefit from visual knowledge: multiple choice question answering tasks including PIQA~\cite{bisk2020piqa} which requires physical commonsense reasoning, CSQA~\cite{talmor2018commonsenseqa} for general understanding of commonsense reasoning, OBQA~\cite{mihaylov2018can} that needs elemenatry-level science knowledge, and RiddleSense (RS)~\cite{lin2021riddlesense} for complex understanding of figurative language, and binary classification task including Visual Paraphrasing (VP)~\cite{lin2015don} that needs scene understanding.
We use in-house test sets made from training sets for PIQA and CSQA since test set is not provided to public.
We list the data statics in Table~\ref{tab:dataset}.
Moreover, We additionally test on GLUE~\cite{wang-etal-2018-glue} to evaluate the general text understanding.

\subsection{Evaluation Protocol}
% \wj{fewshot}
% \xiang{We should have all the details about what settings we're experimenting with the methods; and metrics used in the tasks.}
We evaluate the models in both fully supervised and low-resource settings.
For both settings, we consider accuracy for 5 different classification tasks and get average performance over tasks to check the final performance.
In the fully supervised setting, we evaluate models with 3 different random seeds and report the average accuracy.
In the low-resource setting, we set the size of the train data to 64 or 128. For each experiment, we run over 5 different sub-samples and show the average accuracy.

% randomly sample 64 or 128 examples from train data
% We test models in both low-resource and fully supervised settings.
% In a low-resource setting, we investigate the fast learning and generalizability to new tasks with a few training samples.
% For this, we randomly sample 5 different training sets of 64 and 128 examples from training data and compute average scores..
% For the full training data setup, we train models with 3 different random seeds.

\section{Method}
%In this section, we introduce approaches we will analyze: text-based approaches and vision-based approaches.
In this section, we introduce the following two approaches to integrate visual knowledge into LMs:
(1) \textit{text knowledge transfer}; and (2) \textit{cross-modal knowledge transfer}.
Throughout this section, we assume the data is a collection of image $x^v$ and caption $x^l$ pairs $\left\{(x_i^v, x_i^l)\right\}_{i=1}^{m}$ ($m$ is the size of the pairs) and image encoder $f_V$ and text encoder $f_L$ are given.
Note that we use the same text encoder.
% \xiang{first 3-4 sentences to overview the kinds of methods we consider in this analysis. Provide necessary notations and clarify how we get the initial weights of the text encoder.}

% \subsection{pre-training Objectives}
% \para{Masked language modeling.}
% As an intermediate pre-training on text corpora, we follow BERT~\cite{devlin2018bert} to use masked language modeling (MLM) objective.
% We choose 15\% of the token positions and replace them with \texttt{[MASK]} 80\% of the time, a random token 10\% of the time, the unchanged token 10\% of the time.
% Thus, MLM loss is defined as follows when $i$-th token position is masked.
% \begin{equation}
%     l_{\text{MLM}}(x_i) = - \log p(x_i | x^{\text{masked}}),
% \end{equation}
% where $x_i$ is the $i$-th token and $x^{\text{masked}}$ is a masked text.

% \para{Knowledge distillation.}
% Inspired by VidLanKD~\cite{tang2021vidlankd}, we distill visual knowledge from a pre-trained vision-language model, VL-BERT-large, which is knowledgeable about grounded language.
% Additionally, we distill knowledge from BERT-large to examine whether we can benefit from larger models.
% We do masked language modeling on Wiki103~\cite{merity2016pointer}, a subset of English Wikipedia, in the knowledge distillation step.
% For knowledge distillation, we adopt Neuron Selectivity Transfer (NST)~\cite{huang2017like}, which proves the effectiveness in VidLanKD~\cite{tang2021vidlankd}.

% \para{Contrastive learning.}
% Contrastive learning aims to learn representations by pulling aligned pairs closer and pushing unaligned pairs apart.

\subsection{Text Knowledge Transfer}

% \xiang{
% 0) initial model;
% 1) MLM; 
% 2) SimCSE; SimCSE + MLM
% 3) SupCL.
% }
For text knowledge transfer, we investigate following pre-training objectives:
(1) \textit{masked language modeling}; and (2) \textit{text contrastive learning}.
%We include baselines such as BERT with masked language modeling~\cite{devlin2018bert} and SimCSE~\cite{gao2021simcse}.
% \wj{Other text baselines by Dongho}

\paragraph{Masked Language Modeling (MLM)}
Following BERT~\cite{devlin2018bert}, we select 15\% of input tokens and replace them with \texttt{[MASK]}.
Of the selected tokens, 80\% are replaced, 10\% are not changed and 10\% are replaced by random vocabulary token.
Here, we employ dynamic masking, which performs random masking and replacement during training to prevent the same masking for the same examples~\cite{liu2019roberta}.
MLM objective is the cross-entropy loss for masked token predictions :
\begin{equation}
    \ell_{\text{MLM}}(x_i^l) = - \log p(x_i^l | x^{\text{masked}}),
\end{equation}
where $x_i$ is the $i$-th token and $x^{\text{masked}}$ is a mask.

\paragraph{Text Contrastive Learning (TCL)}
Contrastive learning aims to learn representations by pulling positive pairs closer and pushing negative pairs apart. 
Here, we employ the contrastive framework with cross-entropy objective and in-batch negatives ~\cite{pmlr-v119-chen20j, gao2021simcse}.
Given a text encoder $f_L$, and a caption $x_{i}^l$, we first get text representations using the encoders $h_{i}^{l} = f_L(x_{i}^{l})$.
Following ~\citet{gao2021simcse}, we create identical positive sample ${h}_{i}^{l^{+}}$ by different dropout representations.
The contrastive loss is defined as follows:
\begin{equation}
\ell_{i}^{l} = - \log \frac{e^{\text{sim}({h}_{i}^{l}, {h}_{i}^{l^{+}})/\tau}}{\sum_{j=1}^{N} e^{\text{sim}({h}_{i}^{l}, {h}_{j}^{l})/\tau}},
\label{eq:cl1}
\end{equation}
where $N$ is a batch size and $\text{sim}(\cdot)$ represents cosine similarity, i.e.,  $\text{sim}(u,v)= u \cdot v / \lVert u \rVert \lVert v \rVert$. $\tau$ represents a temperature parameter. 
\subsection{Cross-modal Knowledge Transfer}
\label{sec:cl}

% \xiang{
% Define some key variables and function notations, and use them throughout the introduction here.
% 0) initial model;
% 0) VL transformers are also baseline -- e.g., VL-BERT
% 1) MLM KD from VL model --- formalize the tech details;
% 2) CL (also reference to CLIP and ConVIRT and mention the subtle differences on implementation);
% 3) our implementation of VidLanKD method - re-cap the method with notations; mention differences of the implementation in terms of: 1) data \& video encoding; 2) we try two different losses
% 4) CL w/ LM perturbation;
% 5) CL w/ data augmentation + LM perturbation;
% }

Language models might learn additional information from visual sources such as images and captions.
So we include a variety of vision-based approaches and investigate the approaches whether they can benefit from visual sources.
We introduce vision-based approaches as follows.

% \para{VL-BERT} 
% \wj{pretraining data is different}

\paragraph{Voken Classification}
Vokenization~\cite{tan2020vokenization} employs token-level text-to-image retrieval to transfer visual knowledge.
It aligns language tokens to their related images (called ``vokens'') to transfer visual knowledge into LMs, and call it ``voken classification''.
Given text $x$ and a voken $v_i$ for the $i$-th token, the loss is defined as
\begin{equation}
    \ell_i^{\text{voken}} = - \log(p(v_i|x)).
\end{equation}
% \xiang{mentioned Vokenization in this part?}
% \cz{talk about how you add classification layer here}
Similar to masked language modeling, it classifies each token to a corresponding voken. 
Vokenization trains language models with the voken classification task and MLM.

\begin{figure}[t!]
    \centering 
    \includegraphics[width=0.9\columnwidth]{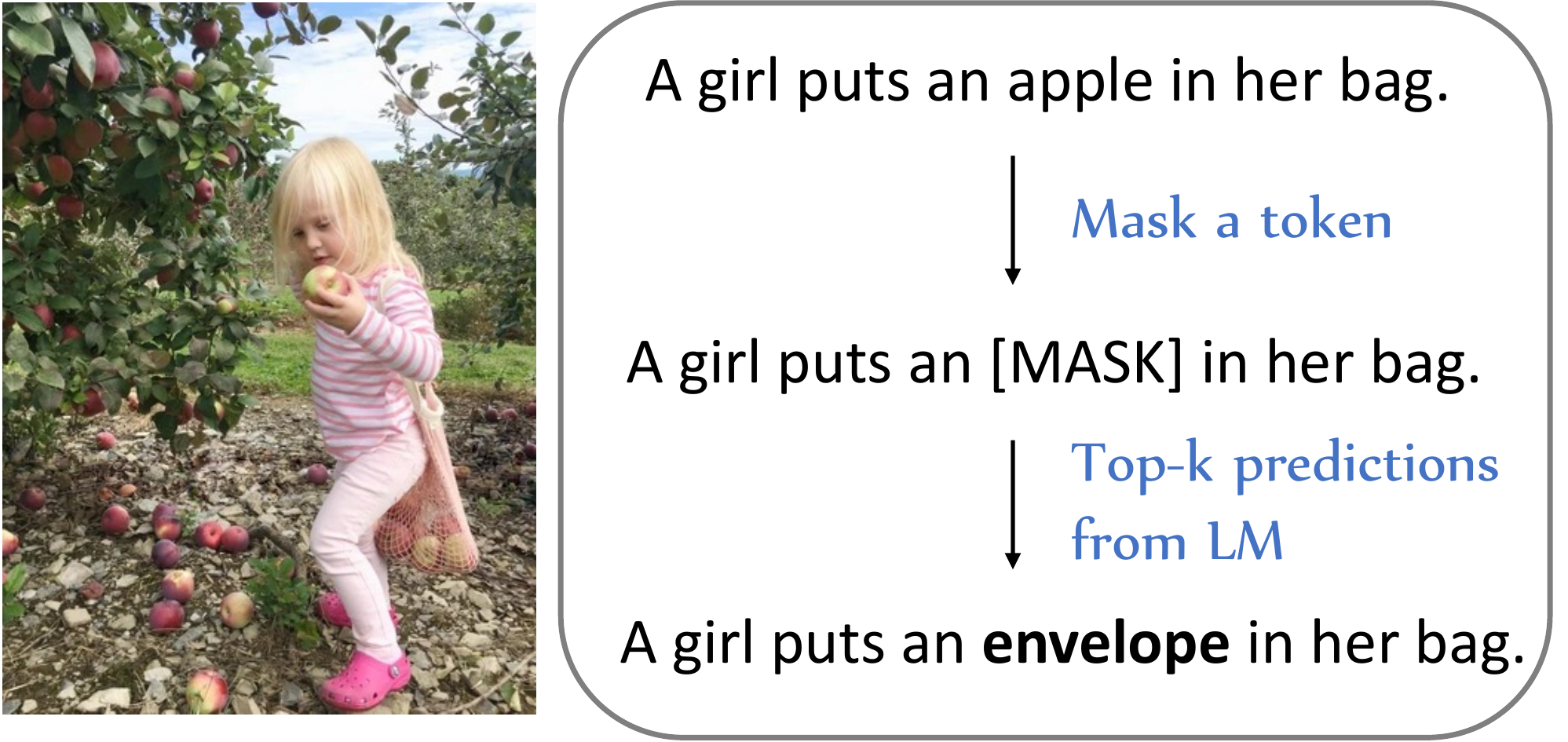}
    \caption{\textbf{LM perturbation.} We create adversarial negatives using language models.}
    \label{fig:perturbation}
\end{figure}

\paragraph{Masked Language Modeling with Visual Clues}
% \xiang{Since VL-BERT can be seen as doing a mixed of intermediate pretraining from BERT; let's move its detailed introduction to here.}
VL-BERT~\cite{su2019vl} adopts masked language modeling with visual clues in which models are given a caption with masked tokens and an image and predict the masked tokens using visual clues.
VL-BERT is pre-trained on Conceptual Captions~\cite{sharma2018conceptual} as an image-caption corpus, and BooksCorpus~\cite{zhu2015aligning} and English Wikipedia as text-only corpora.
It shows its effectiveness in many vision-language tasks.
% We raise a question that can this vision-language model also succeed in NLP tasks? 
We investigate whether this model also succeed in NLP tasks and compare it with others.
% \cz{not sure whether you use VL-BERT here or just as a comparison. In the former case, what is the input, output and loss function here}

% \para{Pre-trained vision-language models.}
% We investigate a pre-trained vision-language model, VL-BERT~\cite{su2019vl}, which succeeded in many multi-modal tasks.
% We raise a question that can this vision-language model also succeed in NLP tasks? 
% VL-BERT is pre-trained on Conceptual Captions~\cite{sharma2018conceptual} as an image-caption corpus, and BooksCorpus~\cite{zhu2015aligning} and English Wikipedia as text-only corpora.
% They include masked language modeling (MLM) as a pre-training task.

\paragraph{Cross-modal Contrastive Learning (CMCL)}
To harness the visual knowledge from image-caption datasets, we adopt contrastive loss on image and text vectors.
% The pre-training task is to predict whether the given images and captions are correctly paired or not.
Given an image encoder $f_V$, a text encoder $f_L$, and an image-caption pair $(x_i^v, x_i^l)$, we first get image and text representations using the encoders $h_i^v = f_V(x_i^v), h_i^l = f_L(x_i^l)$.
Then the contrastive learning objective contains two loss functions: an image-to-text contrastive loss $\ell^{(v,l)}$ and a text-to-image contrastive loss $\ell^{(l,v)}$. The image-to-text contrastive loss is defined as follows:
\begin{equation}
\ell_i^{(v,l)} = - \log \frac{e^{\text{sim}(h_i^v, h_i^l)/\tau}}{\sum_{j=1}^{N} e^{\text{sim}(h_i^v, h_j^l)/\tau}},
\label{eq:cl2}
\end{equation}
where $N$ is a batch size and $\text{sim}(\cdot)$ represents cosine similarity.
This loss encourages a closer distance between representations of aligned image-caption pairs than unaligned pairs given an image and multiple captions.
Similarly, the text-to-image contrastive loss $\ell^{(l,v)}$ is defined as follows:
\begin{equation}
\ell_i^{(l,v)} = - \log \frac{e^{\text{sim}(h_i^l, h_i^v)/\tau}}{\sum_{j=1}^{N} e^{\text{sim}(h_i^l, h_j^v)/\tau}}.
\label{eq:cl3}
\end{equation}
The final loss is defined as
\begin{equation}
    L = \frac{1}{N}\sum_{i=1}^N ( \ell_i^{(v,l)} + \ell_i^{(l,v)}).
    \label{eq:cont}
\end{equation}
CLIP~\cite{radford2021learning} and ConVIRT~\cite{zhang2020contrastive} also adopt contrastive learning, but we freeze the image encoder in training and use the trained text encoder for downstream tasks.

% \para{CL with LM perturbation.}
\paragraph{CMCL with Adversarial Negative Samples (ANS)}
% \xiang{there's more we can write about the tech details of this method. Can reference the text rewriting part in UNIMO paper}
As in-batch negatives in CMCL are not challenging enough for models to distinguish,
% \cz{As in-batch negatives may be easy for the model to distinguish, we present...}
% Here, 
we present adversarial negative sampling strategy to improve CMCL.
Given an image-caption pair $(x_i^v, x_i^l)$, we define a LM-perturbed sentence $x_i^{l^{-}}$, which is a hard negative where $n$ is replaced with a different word $n^{\prime}$ from a probability distribution of PTLMs.
We expect the $l^{-}$ is syntactically correct and plausible sentence even the word $n$ is replaced to $n^{\prime}$, while it does not semantically match to the corresponding image $x_i^v$.
With such hard negative, we try to make more challenging task so that models can effectively learn from the task.
For example, we choose a word `girl' in the sentence `A girl puts an apple in her bag.' in Figure~\ref{fig:perturbation}.
Then we mask the word with \texttt{[MASK]} token to do masked token predictions by PTLMs.
Then we get top-$k$ predictions from language models and replace the masked tokens with one of the predicted ones.
To avoid false negative sentences which may have the same semantics as the original sentence, we introduce an additional filtering step: if the masked predictions are synonyms or hypernyms of the original tokens, we discard the predictions. We use WordNet~\cite{miller1995wordnet} to find synonyms and hypernyms.
The contrastive loss with hard negative is defined as follows:
\begin{equation}
- \log \frac{e^{\text{sim}(h_i^v, h_i^l)/\tau}}{\sum_{j=1}^{N} e^{\text{sim}(h_i^v, h_j^l)/\tau} + \sum_{k=1}^{M} e^{\text{sim}(h_i^v, h_j^{l^{-}})/\tau}},
\end{equation}
where $M$ is the number of hard negative samples per positive pair. This formula is only for image-to-text contrastive loss $\ell^{(v,l)}$ and final loss is defined to same as equation~\eqref{eq:cont}.
% These perturbed sentences are challenging for language models to differentiate since we create them with language models 
\begin{table*}[!t]
	\centering
	\small
	\resizebox{\textwidth}{!}{
		\begin{tabular}{llccccccccccccc}
            \toprule
             & \multirow{2}{*}{\textbf{Model}} & 
             \multicolumn{2}{c}{\textbf{PIQA}} & \multicolumn{2}{c}{\textbf{VP}} & \multicolumn{2}{c}{\textbf{CSQA}} & \multicolumn{2}{c}{\textbf{OBQA}} & \multicolumn{2}{c}{\textbf{RiddleSense}} & \multicolumn{2}{c}{\textbf{Average}} \\
             \cmidrule(lr){3-4} \cmidrule(lr){5-6} \cmidrule(lr){7-8} \cmidrule(lr){9-10} \cmidrule(lr){11-12} \cmidrule(lr){13-14} & & 64 & 128  & 64 & 128  & 64 & 128  & 64 & 128 & 64 & 128    &64 &128  \\
            \midrule
            - & BERT-base   
            &52.6$_{\pm 0.9}$ &53.8$_{\pm 0.1}$ &85.9$_{\pm 1.1}$ &86.6$_{\pm 0.7}$ &35.8$_{\pm 0.7}$ &37.8$_{\pm 0.3}$ &31.3$_{\pm 1.2}$ &32.0$_{\pm 0.7}$ &24.7$_{\pm 0.1}$ &25.2$_{\pm 0.2}$ &46.1 &47.1 \\
            \midrule
            \multirow{4}{*}{\rotatebox{90}{\hspace*{-6pt}Caption}} 
             & MLM 
             & 53.1$_{\pm 0.2}$ & 54.3$_{\pm 0.3}$ & 86.5$_{\pm 0.3}$ & 87.3$_{\pm 0.4}$ & 35.7$_{\pm 0.3}$ & 36.7$_{\pm 0.1}$ & 33.4$_{\pm 0.6}$ & 34.2$_{\pm 0.3}$ & 26.3$_{\pm 0.1}$ & 26.5$_{\pm 0.2}$ & 47.0 & 47.8 \\
             & TCL 
             & 52.6$_{\pm 0.5}$ & 52.9$_{\pm 0.6}$ & 86.4$_{\pm 0.1}$ & 88.0$_{\pm 0.1}$  & 35.7$_{\pm 0.2}$ & 36.1$_{\pm 0.3}$ & \textbf{34.2$_{\pm 1.4}$} & \textbf{35.2$_{\pm 0.7}$} & \textbf{30.3$_{\pm 0.5}$} & \textbf{30.7$_{\pm 0.4}$} &47.8  &48.5 \\
             & TCL + MLM
             & 53.6$_{\pm 0.7}$ & 54.6$_{\pm 0.2}$  & 84.2$_{\pm 0.2}$ & 87.6$_{\pm 0.3}$  & 33.6$_{\pm 2.2}$ & 35.1$_{\pm 0.6}$  & 31.8$_{\pm 2.3}$ & 34.3$_{\pm 0.5}$  & 20.6$_{\pm 0.0}$ & 20.6$_{\pm 0.0}$ & 44.7 & 46.4 \\
             & TCL + ANS 
             & 50.0$_{\pm 0.7}$ & 50.5$_{\pm 0.6}$ & 67.3$_{\pm 0.4}$ & 68.2$_{\pm 0.7}$ & 26.8$_{\pm 1.2}$ & 27.5$_{\pm 0.5}$ & 33.4$_{\pm 1.1}$ & 35.0$_{\pm 1.0}$ & 26.1$_{\pm 1.7}$ & 26.5$_{\pm 1.8}$ & 40.7 & 41.5\\
             & TCL + PSA + ANS
             & 51.1$_{\pm 0.1}$ & 51.2$_{\pm 0.4}$ & 66.0$_{\pm 0.0}$ & 66.0$_{\pm 0.0}$ & 22.7$_{\pm 0.9}$ & 22.9$_{\pm 0.1}$ & 30.2$_{\pm 3.1}$ & 31.8$_{\pm 0.4}$ & 23.5$_{\pm 1.2}$ & 25.2$_{\pm 1.5}$ & 38.7 & 39.4 \\
                %  & KD on BERT-large &54.1 &55.0 &64.5 &73.7 &88.0 &93.6 &38.3 & & &56.7 &57.8 & &39.8 &42.6 & \\
            \midrule
            \multirow{8}{*}{\rotatebox{90}{\hspace*{-6pt}Caption-Image Pairs}} 
             & VL-BERT-base 
             &53.1$_{\pm 0.6}$ &53.9$_{\pm 0.4}$ &\underline{88.5$_{\pm 0.3}$} &88.4$_{\pm 0.5}$ &36.2$_{\pm 0.7}$ &36.8$_{\pm 0.8}$ &33.4$_{\pm 1.2}$ &34.6$_{\pm 1.2}$ &26.1$_{\pm 0.8}$ &26.1$_{\pm 0.9}$ & 47.7 & 48.5\\
            & Vokenization 
            &50.5$_{\pm 0.5}$ &51.1$_{\pm 0.4}$ &68.8$_{\pm 1.6}$ &78.1$_{\pm 1.9}$ &19.2$_{\pm 1.4}$ &21.5$_{\pm 0.8}$ &31.2$_{\pm 2.7}$ &33.2$_{\pm 2.2}$ &17.1$_{\pm 0.5}$ &16.7$_{\pm 0.7}$ & 37.3 & 40.1 \\
            & VidLanKD 
            &55.0$_{\pm 0.4}$ &55.6$_{\pm 0.5}$ &86.7$_{\pm 0.5}$ &\underline{88.5$_{\pm 0.5}$} &37.1$_{\pm 1.0}$ &38.6$_{\pm 0.5}$ &31.8$_{\pm 1.3}$ &32.6$_{\pm 1.0}$ &24.4$_{\pm 0}$ &24.4$_{\pm 0}$ &47.0 &47.9  \\
            & VidLanKD variant 
            &\underline{55.3$_{\pm 0.3}$} &\underline{55.2$_{\pm 0.4}$} &87.4$_{\pm 0.1}$ &88.2$_{\pm 0.6}$  &\underline{37.3$_{\pm 1.2}$} &\underline{38.9$_{\pm 0.5}$} &32.4$_{\pm 2.1}$ &32.2$_{\pm 1.1}$ &24.4$_{\pm 0.0}$ &24.4$_{\pm 0.0}$ & 47.3 & 47.7 \\
            & CMKD (VL-BERT-large) 
            &54.7$_{\pm 0.5}$ &54.5$_{\pm 0.2}$ &86.5$_{\pm 0.8}$ &88.4$_{\pm 0.4}$ &36.7$_{\pm 0.4}$ &38.5$_{\pm 0.4}$ &29.8$_{\pm 0.8}$ &31.7$_{\pm 0.2}$ &25.2$_{\pm 0.1}$ &25.2$_{\pm 0.0}$ & 46.5 & 47.6   \\
           & CMCL 
           &54.7$_{\pm 0.4}$ &55.1$_{\pm 0.1}$ &87.9$_{\pm 0.3}$ &\textbf{88.9$_{\pm 0.2}$} &36.3$_{\pm 0.3}$ &38.4$_{\pm 0.4}$ &31.1$_{\pm 1.1}$ &32.8$_{\pm 0.9}$ &25.0$_{\pm 0.2}$ &25.4$_{\pm 0.4}$ & 47.0 & 48.1\\
           & CMCL + ANS 
           &\textbf{55.4$_{\pm 0.1}$} &\textbf{55.7$_{\pm 0.2}$} &88.1$_{\pm 0.9}$ &\textbf{88.9$_{\pm 0.7}$} &\textbf{37.5$_{\pm 0.8}$} &\textbf{39.0$_{\pm 0.2}$} &32.2$_{\pm 0.7}$ &32.0$_{\pm 0.6}$ &\underline{27.4$_{\pm 0.0}$} &27.5$_{\pm 0.1}$ & \underline{48.1} & \underline{48.6}\\
          & CMCL + PSA + ANS
          &\textbf{55.4$_{\pm 0.2}$} &55.1$_{\pm 0.2}$ & \textbf{88.8$_{\pm 1.0}$} &88.2$_{\pm 0.2}$ &37.0$_{\pm 0.3}$ &38.1$_{\pm 0.3}$ &\underline{34.1$_{\pm 0.4}$} &\underline{34.8$_{\pm 0.9}$} &26.7$_{\pm 0.4}$ &\underline{28.8$_{\pm 0.7}$} & \textbf{48.4} & \textbf{49.0}\\
           \midrule
        \end{tabular}
	}
	\caption{\textbf{Performance (accuracy) in low-resource setting.} We test models on diverse datasets with low-resource learning (64 and 128 training samples). We use captions in the MS COCO dataset for text knowledge transfer methods and images and captions for cross-modal knowledge transfer methods. We get average performance on 64 and 128 training samples. \textbf{Bold} and \underline{underlined} numbers refer to the best and second-best performance, respectively.
% 	\xiang{1) touch on what bold and underscore mean; 2) mention what 64/128 mean; 3) mention the avg columns.}
	}
	\label{tab:mainresults_low}
\end{table*}

% \xiang{As an ablation, can we get a CMCL checkpoint by training BERT from scratch? We can test that on 2-3 datasets and compare with our other CMCL variants for reference.}

\paragraph{CMCL with Positive Sample Augmentation (PSA)}
In ANS, we filter perturbed sentences where the masked predictions are synonyms or hypernyms of the original tokens.
% \xiang{1) introduce what it is and how it was implemented; 2) clarify that ANS is also included in the method}
Instead of excluding these perturbed sentences, another option is to include them as additional positive samples $l^{+}$ to the paired images. We name this as positive sample augmentation (PSA). It also adopts LM-perturbed negative samples as in ANS.
% and thus we augment the training data with additional positive and negative samples.
% \xiang{Can we include notations for the set of synonyms/hyperymns, so we can reference them here. It is unclear in the writing how we implement this variant --- I assume you mean the augmented captions are paired with the image to form additional ``positive img-caption pairs"? Do we still consider ANS in this case? 
% And that also means we're expanding the training instances. These should be made more clear. Also touch on that we need more training steps for this variant because of the expanded training set.}

\paragraph{Cross-modal Knowledge Distillation (CMKD)}
Cross-modal knowledge distillation is to transfer knowledge between different modalities, e.g., image modality and text modality.
In this category, CMKD is to transfer knowledge from a teacher model which is knowledgeable about visual information.
% , VidLanKD~\cite{tang2021vidlankd}
% \xiang{2-3 overview sentences to introduce what this family of methods are doing in general.}
VidLanKD~\cite{tang2021vidlankd} also utilizes a cross-modal knowledge distillation method to help with general language understanding.
A teacher model is first trained using contrastive learning on a video-text dataset, and then it transfers its knowledge to a student language model using KD on a text corpus.
Their contrastive learning loss (hinge loss) is defined as 
\begin{multline}
    L = \sum_i^N [\max(0, \alpha - \text{sim}(h_i^v, h_i^l) + \text{sim}(h_i^{v'}, h_i^l))\\ 
    + \max(0, \alpha - \text{sim}(h_i^v, h_i^l) + \text{sim}(h_i^v, h_i^{l'}))],
    \label{eq:vid}
\end{multline}
where $v'$ and $l'$ are a random image and caption text, respectively. $\alpha$ is the margin between the similarities of a positive pair and a negative pair.
Instead of video datasets, we use a MS COCO dataset to train a teacher model and use two versions of contrastive learning, equations~\eqref{eq:cont} and \eqref{eq:vid}.

% \para{KD on a VL model.}
% \xiang{I move this to here; please adjust and smooth out the transition.}
As another version of CMKD, we consider distilling visual knowledge from a pre-trained vision-language model, VL-BERT, which is knowledgeable about grounded language.
We adopt masked language modeling on Wikitext103~\cite{merity2016pointer}, a subset of English Wikipedia, in the knowledge distillation step.
For knowledge distillation, we adopt Neuron Selectivity Transfer (NST)~\cite{huang2017like}, which proves the effectiveness in VidLanKD~\cite{tang2021vidlankd}.

% \subsection{Other Baselines}

% \para{Pre-trained language models.}
% \xiang{List the bunch of LMs we're comparing}

% \xiang{
% Additional VL methods for comparison:
% 1) VL-BERT-base \& large as text encoder;
% 2) MLM-KD from VL-BERT over Wiki;
% 3) VidLanKD two versions: add VidLanKD original version --> CL w/ some margin loss
% 4) ours w/o perturbed neg examples --> 12k checkpoint --> similar to CLIP/ConVIRT with some subtle differences on implementation
% 5) other variants we tried: (i) ours w/ perturbed negs; (ii) ours w/ perturbed negs + data augmentation}

% We include baselines such as KD~\cite{huang2017like} on VL-BERT~\cite{su2019vl}, and VidLanKD~\cite{tang2021vidlankd}.

\section{Experimental Settings}
% \xiang{Add all details about hps config and search; baseline implementation details; evaluation details.}
For all the approaches, we use \texttt{bert-base-uncased}~\cite{devlin2018bert} as text encoder $f_L$ and ResNeXt101~\cite{xie2017aggregated} as an image encoder $f_V$.
We continue to pre-train the encoders in our experiments.
For text knowledge transfer,
(1) MLM follows the exact setting of codebase in huggingface\footnote{\scriptsize\url{https://github.com/huggingface/transformers/tree/master/examples/pytorch/language-modeling}} which uses dynamic masking strategy to conduct language modeling task.
(2) TCL conducts contrastive learning with $f_L$. We choose the best checkpoint by the best spearman correlation on STSb~\cite{cer-etal-2017-semeval}.
For cross-modal knowledge transfer, 
(1) CMKD explores VL-BERT, Vokenization, and VidLanKD approaches. Here, we use VL-BERT-large model to do CMKD. We use the VL-BERT and Vokenization checkpoints from their official codebases\footnote{\scriptsize\url{https://github.com/jackroos/VL-BERT}, \url{https://github.com/airsplay/vokenization}}. VidLanKD trains a teacher model by two versions of contrastive learning (equations~\eqref{eq:cont} and \eqref{eq:vid}) on MS COCO dataset. We set $\alpha=1$ in VidLanKD (equation~\eqref{eq:vid}).
(2) CMCL conducts contrastive learning with $f_L$ and $f_V$. Here, we set $\tau=0.05$ (equations~\eqref{eq:cl1} and \eqref{eq:cl2}).
(3) CMCL with ANS chooses three noun words or verb words to do masked prediction and use top-5 predictions from $f_L$ as replacement. We filter out synonyms and hypernyms of original words using WordNet~\cite{miller1995wordnet}.
(4) CMCL with PSA includes the perturbed sentences with synonyms and hypernyms as additional positive samples.
In CMCL, we adopt ResNeXt101~\cite{xie2017aggregated} as an image encoder $f_V$ and BERT as a text encoder $f_L$.
% hp
TCL and CMCL train with batch size 64, maximum sequence length 20, learning rate 1e-4 for 3 epochs. For fine-tuning on downstream tasks, we do grid search on learning rates \{5e-5, 1e-4, 3e-4, 4e-4, 5e-4, 6e-4\} and choose the best learning rate.
We set maximum epochs to 30 in low-resource and 15 in fully supervised settings.

% evaluation subsample
% We test models in low-resource learning and fully supervised learning.
% Specifically, we choose 64 and 128 training examples from training data for low-resource learning.
% We randomly sample 5 different training examples and compute average scores.
% For the fully supervised setup, we test models with 3 different random seeds.

\begin{table}[!t]
	\centering
% 	\small
	\resizebox{\columnwidth}{!}{
		\begin{tabular}{llccccccccc}
            \toprule
            &\textbf{Model}  & \textbf{PIQA} & \textbf{VP} & \textbf{CSQA} & \textbf{OBQA} & \textbf{RiddleSense} & \textbf{Average}\\
            \midrule
            - & BERT-base   
            &62.5$_{\pm 1.3}$ &93.1$_{\pm 0.4}$ & 53.2$_{\pm 1.2}$ &52.2$_{\pm 0.5}$ &38.9$_{\pm 0.9}$ & 59.9 \\
            \midrule
            \multirow{4}{*}{\rotatebox{90}{\hspace*{-6pt}Caption}} 
             & MLM 
             & 63.8$_{\pm 0.9}$ & 93.5$_{\pm 0.1}$ & 52.6$_{\pm 0.3}$ & \underline{53.9$_{\pm 1.1}$} & 39.3$_{\pm 1.4}$ & \textbf{60.6} \\
             & TCL 
             & 62.1$_{\pm 0.5}$ & 93.5$_{\pm 0.4}$ & 49.0$_{\pm 0.5}$ & \textbf{54.1$_{\pm 1.0}$} & \textbf{41.2$_{\pm 0.3}$} & \underline{60.1} \\
             & TCL + MLM
             & 62.3$_{\pm 0.7}$ & 93.2$_{\pm 0.3}$ & 49.0$_{\pm 0.4}$ & 49.0$_{\pm 0.8}$ & \underline{40.5$_{\pm 0.5}$} & 58.8\\
             & TCL + ANS 
             & 60.1$_{\pm 1.2}$ & 93.3$_{\pm 0.1}$ & 47.0$_{\pm 0.1}$ & 50.2$_{\pm 0.9}$ & 36.7$_{\pm 0.8}$ & 57.4 \\
             & TCL + PSA + ANS 
             & 59.5$_{\pm 1.0}$ & 92.4$_{\pm 0.3}$ & 34.0$_{\pm 1.3}$ & 44.6$_{\pm 1.4}$ & 28.4$_{\pm 2.3}$ & 51.7\\
                %  & KD on BERT-large &54.1 &55.0 &64.5 &73.7 &88.0 &93.6 &38.3 & & &56.7 &57.8 & &39.8 &42.6 & \\
            \midrule
            \multirow{8}{*}{\rotatebox{90}{\hspace*{-6pt}Caption-Image Pairs}} 
            & VL-BERT-base 
            & 63.8$_{\pm 1.5}$ &93.6$_{\pm 0.1}$ &50.3$_{\pm 1.1}$ & 49.6$_{\pm 2.3}$ &39.1$_{\pm 1.0}$ & 59.2 \\
            & Vokenization 
            & 58.4$_{\pm 5.1}$ &92.7$_{\pm 0.3}$ &45.0$_{\pm 0.2}$ & 48.1$_{\pm 0.8}$ &33.5$_{\pm 0.7}$ & 55.5\\
            & VidLanKD 
            & 63.1$_{\pm 1.1}$ & 93.7$_{\pm 0.4}$ & 52.4$_{\pm 0.8}$ & 50.6$_{\pm 3.9}$ & 39.5$_{\pm 1.7}$ & 59.8 \\
            & VidLanKD variant 
            &\textbf{64.1$_{\pm 0.2}$}  &\underline{93.8$_{\pm 0.3}$} &\textbf{53.6$_{\pm 0.5}$} &47.9$_{\pm 4.3}$ &38.8$_{\pm 2.0}$ & 59.6 \\
            & CMKD (VL-BERT-large) 
            &63.8$_{\pm 0.0}$ &93.7$_{\pm 0.7}$ &\underline{53.3$_{\pm 1.4}$} &48.7$_{\pm 3.0}$ &38.7$_{\pm 0.4}$ & 59.6
            \\
           & CMCL 
           &62.7$_{\pm 0.1}$ &93.3$_{\pm 0.3}$ &50.8$_{\pm 0.9}$ & 52.3$_{\pm 0.7}$ &37.6$_{\pm 1.0}$ & 59.2 \\
           & CMCL + ANS 
            &63.5$_{\pm 0.1}$ &93.3$_{\pm 0.3}$ &50.3$_{\pm 0.1}$ &52.9$_{\pm 0.3}$ & 38.4$_{\pm 0.9}$ & 59.7\\
          & CMCL + PSA + ANS  
          &\underline{63.9$_{\pm 0.5}$}&\textbf{94.3$_{\pm 0.1}$} &50.9$_{\pm 0.3}$ &52.4$_{\pm 1.2}$ &39.0$_{\pm 0.3}$ & \underline{60.1}\\
           \midrule
        \end{tabular}
	}
	\caption{\textbf{Performance (accuracy) in fully supervised setting.} \textbf{Bold} and \underline{underlined} numbers refer to the best and second-best performance, respectively.
% 	\xiang{For naming different KD methods in the tables/figures, I think we just use their original naming: VidLanKD, Vokenization; except for the CMKD (VLBERT-large)}
	}
	\label{tab:mainresults_full}
\end{table}

\begin{table}[!t]
	\centering
	\resizebox{\columnwidth}{!}{
		\begin{tabular}{llcccccccc}
            \toprule
             & \textbf{Model} & \textbf{RTE} & \textbf{MRPC} & \textbf{STS-B} & \textbf{CoLA} & \textbf{SST-2} & \textbf{QNLI} & \textbf{QQP} &  \textbf{Avg.} 
             \\
            \midrule
            - & BERT-base 
            &\bf 70.0 &	\underline{87.9}&	89.1&	57.4&	91.3&	90.4&	89.3& \bf 82.3 \\
            \midrule
            \multirow{4}{*}{\rotatebox{90}{\hspace*{-6pt}Caption}} 
             & MLM 
             & 62.8&	87.0&	89.1&	53.9&\bf	92.6&	91.1&\bf	90.9   & 81.0 \\
             & TCL 
             & 58.4&	83.1&	88.2&	55.5&	\underline{91.9}&\bf	91.4&\bf	90.9   & 79.9\\
             & TCL + MLM 
             & 54.8&	81.6&	87.2&	53.6&	\underline{91.9}&	90.9&	89.2   & 78.5\\
             & TCL + ANS 
             & 56.3&	83.9&	87.0&	51.5&	91.3&	\underline{91.2}&	\underline{89.4}   & 78.6\\
             & TCL + PSA + ANS 
             & 52.3&	75.6&	81.5&	17.4&	90.0&	85.8&	88.2   & 70.1 \\
                %  & KD on BERT-large &54.1 &55.0 &64.5 &73.7 &88.0 &93.6 &38.3 & & &56.7 &57.8 & &39.8 &42.6 & \\
            \midrule
            \multirow{8}{*}{\rotatebox{90}{\hspace*{-6pt}Caption-Image Pairs}} 
           & VL-BERT-base 
           & 57.4&	85.7&	\underline{89.5}&\bf	58.1&	90.6&	89.7&	88.7& 80.0 \\
           & Vokenization 
           &53.0&	87.0&	83.3&	51.3&	91.4&	89.2&	88.5&   77.7 \\
           & VidLanKD
           & 67.5&	87.8&	89.4&	\underline{57.7}&	90.7&	90.3&	88.6&  \underline{81.7} \\
           & VidLanKD variant 
           &68.5&	87.9&\bf	89.7&	54.9&	91.1&	90.5&	88.6   &81.6\\
           & CMKD (VL-BERT-large) 
           &68.5&\bf	88.5&	89.3&	55.4&	90.9&	89.7&	88.6&   81.6 \\
           & CMCL 
           & 63.5&	82.5&	89.5&	51.1&	90.4&	90.0&	88.4    &79.3\\
           & CMCL + ANS 
           & 69.6&	86.8&	89.4&	56.1&	90.7&	90.5&	88.6   &\underline{81.7} \\
           & CMCL + PSA + ANS 
           & \underline{69.8}&	86.2&	89.0&	55.3&	90.4&	90.5&	88.6   &81.6 \\
           \midrule
        \end{tabular}
	}
	\caption{\textbf{Performance (accuracy) on GLUE benchmark.} \textbf{Bold} and \underline{underlined} numbers refer to the best and second-best performance, respectively.
% 	\cz{results are not good, shall we just mention the average number in text, or exclude this subsection?} 
	}
	\label{tab:glue}
\end{table}

% \begin{table}[!t]
% 	\centering
% 	\resizebox{\columnwidth}{!}{
% 		\begin{tabular}{llcccccccc}
%             \toprule
%              & \textbf{Model} & \textbf{RTE} & \textbf{MRPC} & \textbf{STS-B} & \textbf{CoLA} & \textbf{SST-2} & \textbf{QNLI} & \textbf{QQP} &  \textbf{Avg.} 
%              \\
%             \midrule
%             - & BERT-base & 70.0 &	87.9&	89.1&	57.4&	91.3&	90.4&	89.3& 82.3 \\
%             \midrule
%             \multirow{4}{*}{\rotatebox{90}{\hspace*{-6pt}Text}} 
%              & MLM & \\
%              & TCL & \\
%                 %  & KD on BERT-large &54.1 &55.0 &64.5 &73.7 &88.0 &93.6 &38.3 & & &56.7 &57.8 & &39.8 &42.6 & \\
%             \midrule
%             \multirow{8}{*}{\rotatebox{90}{\hspace*{-6pt}Text+Image}} 
%           & VL-BERT-base & \\
%           & CMKD (VL-BERT-large) & 60.29&	80.24&	79.19&	4.38&	82.22&	79.81&	84.35& 67.2 \\
%           & CMCL &54.15&	77.11&	79.29&	0&	82.34&	78.77&	83.85& 65.0 \\
%           & CMCL w/ ANS &47.29&	77.96&	79.84&	9.98&	81.42&	78.75&	83.92& 65.5 \\
%           & CMCL w/ PSA &48.38&	74.8&	79.45&	3.85&	81.88&	78.44&	31.59& 56.9 \\
%           \midrule
%         \end{tabular}
% 	}
% 	\caption{\textbf{Results on large model.} \xiang{placeholder}
% 	}
% 	\label{tab:large}
% \end{table}

\section{Results and Analysis}
\vspace{-0.1cm}
% \xiang{The key analysis questions can be addressed in this subsection first; then we can use other subsection for different ablation study and analysis.}

% \xiang{
% A more general framing of the analysis questions: 
% (Q1): Can \textbf{text} intermediate pre-training help transfer concept-related knowledge to augment text encoders? 
% --> in what settings it works? low-resource / high-resource?
% (Q2): Can \textbf{cross-modal} intermediate pre-training help transfer visual knowledge to augment text encoders?
% (Q3): How does the amount of training data matter? In what settings it works? low-resource / high-resource?
% (A new subsection to focus on ``How", i.e., the pre-training methods: Q1): What intermediate pre-training objectives are effective for cross-modal knowledge transfer?
% (Q2): How can we guide the intermediate pre-training to focus on most challenging cases for the text encoders?
% (Q3): Does model size matter or change the conclusions?
% (A new subsection: Ablation 1): What types of text pre-training corpora are more helpful?;
% (A new subsection: Analysis 1): How do the models perform on general NLU tasks -- GLUE.
% }

We analyze the main results of intermediate pre-training.
Tables~\ref{tab:mainresults_low} and \ref{tab:mainresults_full} show the main results of low-resource learning and fully supervised learning with the MS COCO captioning dataset, respectively.
We train the models with a few training examples, 64 and 128, to understand the better initialization.
We argue that if a model obtains better performance in the low-resource setup, then it is a faster learner and has better generalization on downstream tasks.

\paragraph{Can text intermediate pre-training help improve text encoders? }
%We first analyze the performance of text intermediate pre-training methods.
Text intermediate pre-training using MLM and TCL on a caption corpus improves the performance on downstream tasks in both low-resource and fully supervised settings.
In particular, TCL shows significant improvement on OBQA and RiddleSense over BERT (p-value $< 0.01$).
These results suggest that text intermediate pre-training on visual-related datasets helps performance on commonsense reasoning tasks.

% Interestingly, intermediate pre-training with combined loss of MLM and TCL decrease the performance overall. \xiang{what's the hypothesis for why this happen?}

\paragraph{Can cross-modal intermediate pre-training help transfer visual knowledge to augment text encoders?}
% \xiang{Writing is rough and disconnected below; please improve them to a coherent story, smooth out the transition, and speak out more about the takeaways.}
We observe that cross-modal intermediate pre-training is helpful in both fully supervised and low-resource settings (See Table~\ref{tab:mainresults_low} and ~\ref{tab:mainresults_full}).
Specifically, CMKD with VidLanKD variant outperforms the baseline by 1.6\% point on the PIQA dataset in fully supervised setting.
CMCL also shows its effectiveness. 
However, we could find that it becomes more powerful when equipped with PSA and ANS.
It suggests that data augmentation for positive and negative sampling is an important factor for CMCL.
In low-resource setting, we find that cross-modal knowledge transfer helps better initialization and lets models learn new tasks faster.

% In the fully supervised setting, we observe that using external knowledge source improves the performance of cross-modal intermediate pre-training in many cases. 
% For example, CMKD with VidLanKD variant outperforms BERT-base by 
% Also, CMCL with PSA and ANS improves CMCL with ANS, suggesting that including additional positive samples is effective to transfer visual knowledge. 

% \wj{what if we have small set of training data? Analysis section. faster learner. prior knowledge. }
% \para{??? low-resource learning.}
% \xiang{Associate the discussion here with ``faster learner/generalization in low-resource settings"}
% In Table~\ref{tab:mainresults_low} of low-resource learning, we observe that using external knowledge, MS COCO data, helps better initialization overall so that models can learn new tasks faster.
% Among various cross-modal knowledge transfer methods, CMCL with PSA and ANS notably improves low-resource learning indicating that it helps model generalize to new tasks.

% For example, intermediate pre-training with MLM on caption text outperforms than BERT-base on all downstream datasets, which suggesting that it helps better initialization.
% In addition, CMCL with ANS also helps better initialization on downstream tasks so it shows the improvement over BERT-base.

\paragraph{What intermediate pre-training objectives are effective for cross-modal knowledge transfer?}
Among various cross-modal knowledge transfer methods, we study which method is the most effective for cross-modal knowledge transfer.
Overall, CMCL with PSA and ANS shows the best performance among all cross-modal methods.
Interestingly, VL-BERT also shows better performance than BERT-base on all datasets in the low-resource setting.
This suggests that exploiting images in masked language modeling task help transfer the knowledge to language models.

% \para{Model size??}
% \xiang{A few senetnces on large model results can go into this part as a paragraph -- main msg is whether the findings are consistent across base/large models.}

\paragraph{What types of knowledge sources are most helpful?}
Here, we investigate whether using an image source in addition to a text source can further improve the model.
To answer this question, we analyze methods from different types of sources: text-only and text-image pair sources.
We focus on the methods that use the contrastive learning objective: TCL and CMCL. 
Note that these two methods share the same objective but CMCL trains on cross modalities which are images and captions while TCL only trains on captions.
Overall, TCL performs slightly better than CMCL in low-resource and fully supervised settings.
Interestingly, additional negative samples (ANS) and positive samples in TCL decreases the performance while they help CMCL to improve the performance.
We conjecture that perturbed sentences in ANS might not be semantically negative to the original sentence so models learn from wrong labels.
% adding additional negative to TCL might confuse the model to sort out perturbed sentences since the perturbed sentences are created from language models \wj{....}

\subsection{Ablation Study}
\vspace{-0.1cm}

\paragraph{How do models perform on general NLU tasks?}
Table~\ref{tab:glue} presents results on GLUE benchmark. 
In GLUE, text intermediate pre-training methods slightly underperform the original BERT-base.
We conjecture that the intermediate pre-training on caption data might sacrifice knowledge of general language understanding.

% \para{Vocabulary overlap.}

\paragraph{Analysis on diverse text corpora}
Table~\ref{tab:corpora} represents text approaches with different pre-training corpora: MS COCO captions~\cite{lin2014microsoft,chen2015microsoft}, GenericsKB~\cite{Bhakthavatsalam2020GenericsKBAK}, BooksCorpus~\cite{Zhu_2015_ICCV}, and WikiText103~\cite{merity2016pointer}.
We sample 250k sentences from each corpus for a fair comparison.
We notice that caption datasets are useful on OBQA and RiddleSense datasets while GenericsKB are the most helpful on PIQA datasets.
Results are expected since GenericsKB contains a lot of everyday statements that contain various types of commonsense.

\begin{table*}[!t]
	\centering
	\small
	\resizebox{\textwidth}{!}{
		\begin{tabular}{llccccccccccccccc}
            \toprule
             & \multirow{2}{*}{\textbf{Model}} & 
             \multicolumn{3}{c}{\textbf{PIQA}} & \multicolumn{3}{c}{\textbf{VP}} & \multicolumn{3}{c}{\textbf{CSQA}} & \multicolumn{3}{c}{\textbf{OBQA}} & \multicolumn{3}{c}{\textbf{RiddleSense}} \\
             \cmidrule(lr){3-5} \cmidrule(lr){6-8} \cmidrule(lr){9-11} \cmidrule(lr){12-14} \cmidrule(lr){15-17} & & 64 & 128 & Full & 64 & 128 & Full & 64 & 128 & Full & 64 & 128 & Full & 64 & 128 & Full \\
            \midrule
            - & BERT-base &52.6$_{\pm 0.9}$ &53.8$_{\pm 0.1}$ &62.5$_{\pm 1.3}$ &85.9$_{\pm 1.1}$ &86.6$_{\pm 0.7}$ &93.1$_{\pm 0.4}$ &35.8$_{\pm 0.7}$ &37.8$_{\pm 0.3}$ & \bf 53.2$_{\pm 1.2}$ &31.3$_{\pm 1.2}$ &32.0$_{\pm 0.7}$ &52.2$_{\pm 0.5}$ &24.7$_{\pm 0.1}$ &25.2$_{\pm 0.2}$ &38.9$_{\pm 0.9}$ \\
            \midrule
            \multirow{2}{*}{\rotatebox{90}{\hspace*{0pt}CP.}} 
             & MLM & 53.1$_{\pm 0.2}$ & \underline{54.3$_{\pm 0.3}$} & 63.8$_{\pm 0.9}$ & 86.5$_{\pm 0.3}$ & 87.3$_{\pm 0.4}$ & \bf 93.5$_{\pm 0.1}$ & 35.7$_{\pm 0.3}$ & 37.7$_{\pm 0.1}$ & \underline{52.6$_{\pm 0.3}$} & 33.4$_{\pm 0.6}$ & \underline{34.2$_{\pm 0.3}$} & \underline{53.9$_{\pm 1.1}$} & 26.3$_{\pm 0.1}$ & 26.5$_{\pm 0.2}$ & 39.3$_{\pm 1.4}$ \\
             
             & TCL & 52.6$_{\pm 0.5}$ & 52.9$_{\pm 0.6}$ & 62.1$_{\pm 0.5}$ & 86.4$_{\pm 0.1}$ & 88.0$_{\pm 0.1}$ & \bf 93.5$_{\pm 0.4}$ & 35.7$_{\pm 0.2}$ & 36.1$_{\pm 0.3}$ & 49.0$_{\pm 0.5}$ & \bf 34.2$_{\pm 1.4}$ & \textbf{35.2$_{\pm 0.7}$} & \textbf{54.1$_{\pm 1.0}$} & \textbf{30.3$_{\pm 0.5}$} & \textbf{30.7$_{\pm 0.4}$} & \textbf{41.2$_{\pm 0.3}$} \\
         \midrule
         \multirow{2}{*}{\rotatebox{90}{\hspace*{0pt}GK.}} 
         & MLM & 53.2$_{\pm 0.1}$ & 53.6$_{\pm 0.4}$ & \bf 64.9$_{\pm 0.1}$ & 86.2$_{\pm 0.9}$ & 87.6$_{\pm 0.3}$ & 93.0$_{\pm 0.3}$ & 34.6$_{\pm 0.7}$ & 35.3$_{\pm 1.3}$ & 51.6$_{\pm 0.5}$ & 31.7$_{\pm 0.9}$ & 32.3$_{\pm 1.0}$ & 53.1$_{\pm 0.9}$ & 25.8$_{\pm 0.6}$ & 26.3$_{\pm 0.1}$ & 39.3$_{\pm 0.7}$ \\
         & TCL & \bf 56.0$_{\pm 1.0}$ & \bf 56.4$_{\pm 0.2}$ & \underline{64.4$_{\pm 0.1}$} & \bf 88.9$_{\pm 0.7}$ & \underline{89.4$_{\pm 0.2}$} & \underline{93.3$_{\pm 0.5}$} & \underline{37.8$_{\pm 1.2}$} & \bf 38.7$_{\pm 0.5}$ & 51.0$_{\pm 0.5}$ & 31.7$_{\pm 0.9}$ & 32.3$_{\pm 1.0}$ & 52.6$_{\pm 0.8}$ & 27.4$_{\pm 0.2}$ & 28.1$_{\pm 0.7}$ & 40.9$_{\pm 0.8}$  \\
         \midrule
          \multirow{2}{*}{\rotatebox{90}{\hspace*{0pt}BC.}} 
         & MLM & \underline{54.1$_{\pm 0.3}$} & 54.1$_{\pm 0.8}$ & 63.3$_{\pm 0.6}$ & 86.4$_{\pm 0.8}$ & 87.5$_{\pm 0.5}$ & 93.0$_{\pm 0.3}$ & 29.8$_{\pm 0.8}$ & 32.1$_{\pm 0.9}$ & 50.8$_{\pm 0.3}$ & 29.6$_{\pm 0.8}$ & 31.4$_{\pm 0.7}$ & 50.2$_{\pm 0.4}$ & 22.6$_{\pm 0.0}$ & 22.7$_{\pm 0.0}$ & 36.7$_{\pm 1.3}$ \\
         & TCL & 52.4$_{\pm 0.1}$ & 53.1$_{\pm 0.4}$ & 63.1$_{\pm 0.3}$ & \underline{87.1$_{\pm 1.9}$} & \bf89.7$_{\pm 0.1}$ & 93.2$_{\pm 0.2}$ & \bf 38.0$_{\pm 0.5}$ & \underline{38.1$_{\pm 1.1}$} & 51.5$_{\pm 0.1}$ & \underline{33.8$_{\pm 2.7}$} & 34.0 $_{\pm 2.1}$ & 55.6$_{\pm 0.4}$ & \underline{28.9$_{\pm 0.4}$} & \underline{29.1$_{\pm 0.3}$} & \bf 41.2$_{\pm 2.3}$\\
         \midrule
      \multirow{2}{*}{\rotatebox{90}{\hspace*{0pt}WT.}}
         & MLM & 52.7$_{\pm 0.2}$ & 53.0$_{\pm 0.3}$ & 63.8$_{\pm 0.6}$ & 85.3$_{\pm 2.8}$ & 88.1$_{\pm 0.3}$ & \bf 93.5$_{\pm 0.1}$ & 33.2$_{\pm 1.4}$ & 34.6$_{\pm 0.5}$ & 52.5$_{\pm 0.2}$ & 32.4$_{\pm 2.3}$ & 33.0$_{\pm 0.7}$ & 52.3$_{\pm 0.3}$ & 24.4$_{\pm 0.0}$ & 24.4$_{\pm 0.0}$ & \underline{39.4$_{\pm 2.0}$}  \\
         & TCL & 52.9$_{\pm 0.9}$ & 53.4$_{\pm 0.4}$ & 62.7$_{\pm 0.6}$ & 67.3$_{\pm 0.6}$ & 68.6$_{\pm 0.7}$ & \underline{93.3$_{\pm 0.3}$} & 31.3$_{\pm 1.6}$ & 32.4$_{\pm 0.7}$ & 48.2$_{\pm 0.3}$ & 31.5$_{\pm 3.5}$ & 33.1$_{\pm 0.6}$ & 53.0$_{\pm 0.0}$ & 24.8$_{\pm 1.3}$ & 24.8$_{\pm 0.6}$ & 36.3$_{\pm 1.0}$ \\
         \midrule
        \end{tabular}
	}
	\caption{\textbf{Results of text knowledge transfer methods with different corpora.} We pre-train text knowledge transfer methods, MLM ans TCL, with different corpora.
	CP is MS COCO captions, GK is GenericsKB, BC is BooksCorpus, and WT is WikiText. \textbf{Bold} and \underline{underlined} numbers refer to the best and second-best performance, respectively.
	}
	\label{tab:corpora}
\end{table*}

\begin{figure}[t!]
\vspace{-0.2cm}
    \centering 
    \resizebox{0.7\columnwidth}{!}{
    \includegraphics{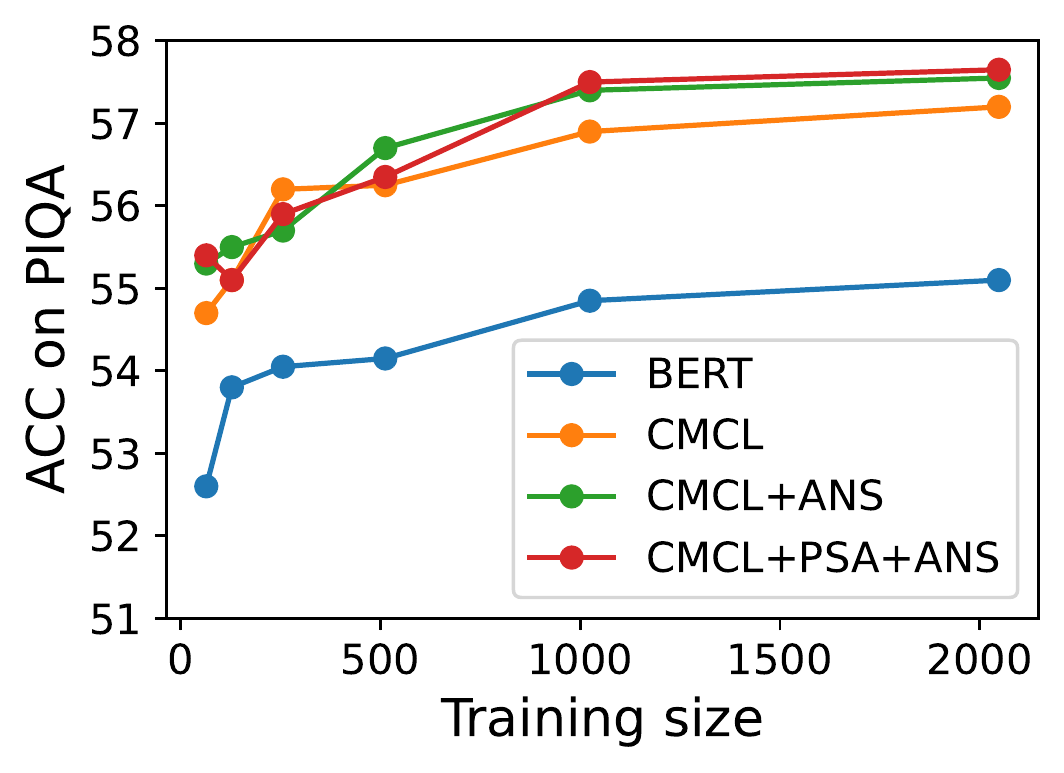}
    }
    \vspace{-0.2cm}
    \caption{\textbf{Results on varying training sizes.} We test methods with different training sizes.}
    \label{fig:trainsize}
\vspace{-0.2cm}
\end{figure}

\paragraph{Different training sizes.}
We test different training sizes on PIQA in Fig.~\ref{fig:trainsize}.
In the experiment, we observe that CMCL consistently outperforms BERT on all training sizes.
Additional negative sample (ANS) improves the CMCL on different training sizes, and positive sample augmentation boosts the performance of CMCL further.
This suggests including perturbed sentences as positive and negative samples are useful to cross-modal knowledge transfer.

% \subsection{Q4: How can we guide the intermediate pre-training to focus on most challenging cases for the text encoders?}
% \wj{results with ANS, PSA compared to CMCL}

% \para{Text baselines.}
% \para{VL baselines.}
% \subsection{Performance over Different Training Sizes}

% \begin{figure}[tb!]
%     \centering
%     {\includegraphics[width=0.8\columnwidth]{figures/trainsize.pdf}}
%     \caption{\textbf{Performance over different training sizes.} 
%     }
%     \label{fig:obj}
% \end{figure}

% \begin{figure}[t!]
%     \centering 
%     \resizebox{0.98\columnwidth}{!}{
%     \includegraphics{figures/reporting_bias.pdf}
%     }
%     \caption{\textbf{Vocabulary overlap between MS COCO and downstream datasets.} We collect top 1k most frequent words from each dataset and we compute the intersection size over the entire vocabulary size.}
%     \label{fig:wordoverlap}
% \end{figure}

\section{Related Work}

\vspace{-0.1cm}
\paragraph{Text Knowledge enhanced methods.}
Recently, huge efforts on integrating knowledge into PTLMs have been made.
One typical form of knowledge is a knowledge graph.
There have been efforts of using knowledge graph to inject entity and relation representations, which are pre-computed from external source, into PTLMs~\cite{zhang-etal-2019-ernie, xu2021does,peters-etal-2019-knowledge, he-etal-2020-bert, dekcor}.
Some other works try to retrieve or generate the sub-graph from the graph to solve the problem~\cite{lin-etal-2019-kagnet,wang-etal-2020-connecting}.
Another existing form of knowledge is extra large-scale corpus.
Works that use such corpus present knowledge-related pre-training objectives such as concept order recovering~\cite{zhou2021pretraining}, entity category prediction~\cite{yu2020jaket} and source of knowledge prediction~\cite{wang-etal-2021-k,calixto2021wikipedia}.
They are mostly focused on injecting world knowledge presented in text, rather than physical and visual commonsense knowledge that can be found in images.

\paragraph{Cross-modal knowledge enhanced methods.}
There is a extensive line of works for a variety of vision-language tasks, such as VL-BERT~\cite{su2019vl}, VisualBert~\cite{li2019visualbert}, and Uniter~\cite{chen2020uniter}.
These models aim to improve vision-language tasks, e.g., VQA~\cite{goyal2017making} and event understanding~\cite{li2022clip}, and they are found to be not effective in improving language tasks~\cite{tan2020vokenization}.
Another line of works is to transfer visual knowledge to language models: Vokenization~\cite{tan2020vokenization} and VidLanKD~\cite{tang2021vidlankd}.
Vokenization employs token-level text-to-image retrieval to transfer visual knowledge to language models. For this, Vokenization introduces 30k vokens and matches each token into the limited voken space. %; it may have approximation errors.
VidLanKD adopts contrastive learning to train a teacher model on video datasets and uses distillation approaches to distill visual knowledge from the teacher to a student model.
% In this study, we investigate these approaches on diverse datasets and examine how to help transferring visual knowledge into language models.  

% \xiang{couple lines of related work. Let's use one bold paragraph title for each: 
% 1) injecting KG info to the LMs (cite some papers) -- but the KG coverage, esp on visual knowledge, is a bottleneck; 
% 2) retrieval-augmented and generation-augmented methods (cite, such as Peifeng's PathGen and Bill's KagNet) -- but it also require identifying a corpus to use, which may not represent visual knowledge well due to reporting biases. 
% 3) an extensive line of work doing VL pretraining for VL transformers. However, these work aim to improve VL tasks and found to be not effective in improving language tasks (can cite VL-BERT or other papers that found this.)
% 4) some recent work like Vokenization, VidLanKD~\cite{tang2021vidlankd} -- make sure to clearly stress on their limitations and our distinctions.
% }

\section{Conclusion}

We study whether intermediate pre-training on visual knowledge can help transfer visual knowledge into LMs.
We investigate text knowledge transfer and cross-modal knowledge transfer using images and captions.
In our empirical analysis, we observe that intermediate pre-training on captions can help improving performance and cross-modal knowledge transfer approaches consistently improve performance.
When the transfer methods are equipped with additional positive and negative samples, they show better performance.
Future works include improving both commonsense reasoning and general language understanding.

% \section*{Acknowledgement}
% \input{070acknowledgement}

\bibliography{bibtex}
\bibliographystyle{acl_natbib}

% \appendix

\clearpage
\appendix
\section{Dataset Properties}

% \xiang{move below to appendix; created a compressed version to summarize (also shown in table?):
% 1) what are the nature and format of these datasets;
% 2) what capability these datasets are testing for models;
% 3) why do we select / focus on these couple datasets;
% 4) what do we want to analyze by including GLUE?
% }
PIQA is a multiple-choice question answering task, which chooses the most appropriate solution for physical commonsense questions, which may need illustration or description of physical interaction in the real world.
VP is to tell if two descriptions are describing the same scene or two different scenes.
While they seem like purely textual tasks, they require visual common sense to answer.
CSQA is a multiple-choice question answering task that requires commonsense reasoning to answer. It is built from ConceptNet~\cite{conceptnet}.
OBQA is a multiple-choice question answering task, which is modeled after open book exams on elementary-level core science questions. The task generally requires open book fact but also additional commonsense which can be learnt from scientific illustration.
RiddleSense is a multiple-choice riddle-style question answering which requires complex commonsense reasoning ability and understanding of figurative language which may benefit from visual knowledge.

\end{document}